\documentclass[wcp]{jmlr}
\usepackage{graphicx}

\jmlrvolume{1}
\jmlryear{2018}
\jmlrworkshop{ICML 2018 AutoML Workshop}

\title{Neural Network Conversion of Machine Learning Pipelines}

 \author{\Name{Man-Ling Sung} \Email{sammi.sung@raytheon.com}\\
   \Name{Jan Silovsky} \Email{jan.silovsky@raytheon.com}\\
 \Name{Man-hung Siu} \Email{man-hung.siu@raytheon.com}\\
   \Name{Herbert Gish} \Email{herb.gish@raytheon.com}\\
 \Name{Chinnu Pittapally} \Email{chinnu.pittapally@raytheon.com}\\
 \addr Raytheon BBN Technologies, 10 Moulton Street, Cambridge, MA 02138}

\begin{document}

\maketitle

\begin{abstract}
Transfer learning and knowledge distillation has recently gained a lot of attention in the deep learning community.  One transfer approach, the student-teacher learning, has been shown  to successfully create ``small'' student neural networks that mimic the performance of a much bigger and more complex ``teacher'' networks. In this paper, we investigate an extension to this approach and transfer from a non-neural-based machine learning pipeline as teacher to a neural network (NN) student, which would allow for joint optimization of the various pipeline components and a single unified inference engine for multiple ML tasks.  In particular, we explore replacing the random forest classifier by transfer learning to a student NN.  We experimented with various NN topologies on 100 OpenML tasks in which random forest has been one of the best solutions.  Our results show that for the majority of the tasks, the student NN can indeed mimic the teacher if one can select the right NN hyper-parameters.  We also investigated the use of random forest for selecting the right NN hyper-parameters.  
\end{abstract}

\section{Introduction}
\label{sec:intro}
Our goal is to explore the use of neural networks as replacements for ML pipeline, or a portion of these pipeline. We will accomplish this replacement by having the neural network (NN) learning from the original pipeline.  This goes beyond the more usual motivation in employing the student teacher paradigm, which goes from large NN to smaller, more deployable ones, e.g., \cite{ba2014deep, Hinton2015}.  While smaller and more deployable are welcome characteristics, we have additional considerations. One consideration is that our converted components may be part of a larger network, and chaining various converted components to form a larger neural network will simplify the joint optimization of all parts of our system.

Mapping various systems into neural networks can have several additional benefits. Specialized hardware, such as GPUs can enhance performance and a neural network may have better generalization performance than the original systems.  Moreover, we expect our systems to operate in dynamic environments and having a unified approach to these changes can enhance the capabilities of these more deployable systems. In particular we can use standard methods for regularizing these networks, which ties in with generalization capabilities as well as methods for adapting the networks to changing conditions.

In this work, our focus is on converting other classifiers into NN classifiers. We understand that a neural network is not necessarily the best classifier in all situations, especially in small data problems. However, our goal is not to exceed the performance of the teacher but rather attempt to match it.  In some situations it may be necessary to rely mainly on the function approximation capabilities of neural networks and at other times we may need to train the neural network using methods of data augmentation. By data augmentation we envision modeling of the feature space based on the original training data and generating new samples based on this model. The samples, coupled with the labels provided by the teacher provide additional training for the neural network.

In some cases, the teacher brings more to the knowledge transfer problem than just generating classification responses. In some particular cases, we can have knowledge of the decision boundary in some form as well as metadata regarding the structure of the particular classifier. For example, for random forest classifiers it has been shown (\cite{Sethi1990},\cite{Biau2016}) that they can be restructured as multi-layered neural networks.  Additionally, \cite{Wang2017} shows how to create random forest inspired neural network architectures.  In our current work, these characteristics are not exploited.

In the following, we discuss in greater detail the student-teacher approach, which for us is conversion to a neural network, followed by a discussion of our experimental results.


\section{Neural Network Conversion}
\subsection{Student-Teacher Knowledge Distillation Learning}
\label{sec:stk}
The goal of knowledge distillation is to transfer knowledge acquired by "teacher" to a "student" such that the student can perform as well or better than the teacher.  Typically, the teacher is a complex system either with large number of parameters and/or an ensemble of classifiers while the student is relatively smaller to enable efficient inference, e.g. in ~\cite{Hinton2015}. For Inductive Transfer Learning defined in \cite{pan2010survey}, consider training data $T= \{(x_1,y_1), \ldots,(x_n,y_n)\}$ where $x$'s are the input features with corresponding labels $y$'s.  The teacher model $M$ is trained using $T$. $M$ can generate the labels of a new training set $T' = \{(x'_1,\hat{y}_1), \ldots, (x'_n,\hat{y}_n)\}$ where $\{\hat{y}\}$ is the set of label posteriors generated by $M$.  $\{x'\}$ in $T'$ and $\{x\}$ in $T$ can be different. In~\cite{Li2014}, $\{x'\}$ includes additional unlabeled data and in~\cite{Cui2017}, $\{x'\}$ and $\{x\}$ are generated by different feature extractors.
\subsection{Student-Teacher over Different System Types}
In typical distillation framework, both teacher and student are neural networks, or as in~\cite{Yu2017}, transfer across very specific systems is possible by tapping into the internal states of the teacher systems. However, the student-teacher formulation in Section~\ref{sec:stk} can be generalized to distill between two different system types with the following considerations.
\begin{enumerate}
\item{Trainability}: The student system can be trained using label posteriors $\hat{y}$.
\item{Feature Handling}: The student system can process the type of input feature $x$.
\item{Student Complexity}: The student system should have enough capacity to learn the decision boundaries of the teacher system.  For example, using a linear classifier as a student will not be able to mimic the classification decision of a deep neural network classifier.
\end{enumerate}
Other than the particular type of student system (i.e. neural networks or random forests), the hyper-parameters of the student and the amount of available training can have significant impact on distillation effectiveness.

\subsection{Training Data for Student}
The quality of the transfer depends heavily on the amount of training data available and the complexity of the student model.  As noted above, the student can be trained with a different data set from those used for training the teacher. While it can be difficult or expensive to obtain manually annotated data, the annotation needed for student training, $\{\hat{y}\}$, can easily be generated using the teacher model.  Thus, extending $T'$ only involves obtaining more $x'$.  This can be accomplished by 1) collecting more unlabeled data which is feasible for many problems; 2) Using $T$ to estimate the input feature distribution, $P(x)$, and then sampling from it. Where $P(x)$ can be estimated using either parametric models, such as GMM, or non-parametric models, such as KNN or any kernel-based distribution estimators; 3) Assuming $P(x)$ to be a uniform distribution and sampling from it. Such $P(x)$ can be suboptimal as discussed in~\cite{Scholkopf2012} but can be useful as a smoothing function. 
\subsection{Initial Approach}
In this paper we focus on a set of random forest teachers and our ability to match the random forest performance with NN classifiers.  We selected random forest classifiers based on their reputation for providing the best performance on a wide range of problems and also their widespread use.  We perform this exploration on a standard set of problems provided by OpenML.  In addition to student-teacher performance comparisons, we also investigate ways to determine the best choice of neural network architecture and hyper-parameters to employ on particular problems.

\section{Experiments}
\label{sec:expt}
\subsection{OpenML}
\label{ssec:openml}
OpenML (Open Machine Learning), founded by \cite{vanschoren2014openml}, is a platform for sharing datasets, ready-to-use models, and problems in machine learning. It provides cross-language APIs that facilitates the reproduction and comparison of different machine learning architectures. 
There are 4 main organization groups, 1. \textbf{Data:} collection of data sets available for definition of ML problems; 2. \textbf{Task:} a formulation of a ML problem and specification of evaluation criteria; 3. \textbf{Flow:} describes a particular solution as a \textit{composition} of primitives/modules performing various tasks - e.g. feature extraction, normalization, classification, etc.
4. \textbf{Run:} describes particular \textit{configuration} of a flow, most importantly, hyperparameters of individual primitives. Hence, multiple Runs can be associated with identical Flow and yield different performance.

\subsection{Experimental setup}
\label{ssec:expt_setup}
First, we identified a Flow, \cite{Randal2017openml}, employing random forest as the backend classifier, which was evaluated for many tasks. The flow we found was composed of three sklearn primitives: preprocessing.imputation.Imputer, decomposition.pca.PCA
and ensemble.forest.RandomForestClassifier. Next, we selected 100 Tasks based on best Runs.

A Student system was built simply by substituting the random forest (RF) classifier by Multi-Layer Perceptron (MLP)\footnote{We relied on sklearn's implementation of MLP classifier \cite{scikit-learn}}. 
For each Task, we used identical set of 600 different configurations of MLPs acting as different Students.
Table~\ref{tab:MLP_Students} tabulates the parameters modified in our configurations.  For parameters not listed, sklearn's MLP defaults are used.
By bottleneck, we refer to the middle layer in systems having 3 or more layers and the relative size presented in Table~\ref{tab:MLP_Students} is relative to the standard layer size in the networks\footnote{A NN with 3 layers, 100 nodes per layer and relative bottleneck size of 0.5, has (100,50,100) nodes in its hidden layers}.

\begin{table}[h]
\begin{center}
    \begin{tabular}{ | l | l | l | l | l |}
    \hline
    Layers & Nodes in layer & Rel. bottleneck size & Activation & Init. learning rate \\ \hline
    1,2,3,4,5 & 10,25,100,200,400 & 0.2, 0.5, 1.0 & relu, tanh & 1e-2,1e-3,1e-4,1e-5 \\ \hline
    \end{tabular}
\end{center}
\caption{Overview of different configurations of MLP Students}
\label{tab:MLP_Students}
\end{table}

\subsection{Student-Teacher Knowledge Transfer}
\label{ssec:expt_ST}
In this study, we focused on knowledge transfer using the original training inputs. Thus, the Student model $M'$ is trained with training data $T' = \{(x_1,\hat{y}_1), \ldots,(x_n,\hat{y}_n)\}$, where $\hat{y} = M(x)$.

The OpenML experiments are designed as 10-fold cross-validation and we followed this experimental setup. This means that for each task, 10 different RF Teachers were trained and the knowledge transfer applied independently for 10 MLP Students with a particular configuration (one of the 600). The final task accuracy is then simply an average over the 10 folds.

\begin{figure}
\centering     
\includegraphics[width=\linewidth]{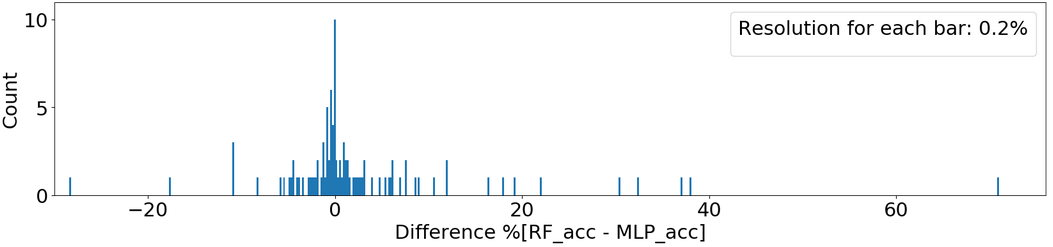}
\caption{Histogram on differences between RF and MLP accuracies on 100 tasks}
\label{fig:RF_diff}
\end{figure}

Fig.~\ref{fig:RF_diff} illustrates the performance difference of the random forest Teachers and the MLP Students. The best performing MLP configuration is considered for each task. Over all tasks, 55\% of Students perform equally well or better than Teacher. On average, the performance of Students is worse by 2.66\%. In terms of the median, the Students perform as well as the Teachers (0.01\% better). The shift between the average and median is caused by few outliers as shown in the right side of the figure. We plan to further investigate why MLP performs so poorly on the few outliers.

For some tasks, the Student surprisingly outperforms the Teacher by a larger margin.
We attribute this partly to natural statistical variations, and partly to the fact that RF partitions the feature space in rectangular regions while MLP has smoother decision boundary which may fit certain problems better.


\subsection{Student Versatility and Complementarity}
\label{ssec:expt_SVC}

Having large number of MLP student configurations (hereafter, we refer to these student configurations as {\em Students}) is impractical and we expect many to have similar performance across Tasks. It is desirable to keep only a smaller set of complementary Students, i.e. student configurations with high performance across many Tasks.

Fig.~\ref{fig:Auto_RF} depicts how varying the number of Student candidates affect the performance across Tasks.
Candidate sets of each size were formed by removing the systems with least contribution to the overall performance.

We found that the single best system\footnote{A DNN with two hidden layers (400,400), relu activation function and initial learning rate of 1e-2} turns out to be very versatile across Tasks as it performs only 0.9\% worse on average compared to the choice of the best Student out of the full inventory of 600 Students.
However, as shown in the figure, picking from 20 Students reduces the gap by half to 0.45\%.

\subsection{Automatic Student Selection}
\label{ssec:expt_auto}

While we can rely on cross-validation experiments to select the best Student, it may still not be feasible to train multiple Students in some practical applications.
Ideally we would be able to automatically select the best Student candidate based on characteristics of the Data, Task and the Teacher.

We carried out a set of experiments using random forest for selecting the best student candidate. Intuitively, the complexity of selecting the best Student grows with the number of Students candidates and the complexity is further accentuated by the limited number of training samples (100 samples corresponding to the 100 Tasks). The RF system for automatic Student selection was trained with a 10-fold cross-validation over the Tasks.

\begin{figure}
  \begin{center}
  \includegraphics[width=0.65\linewidth]{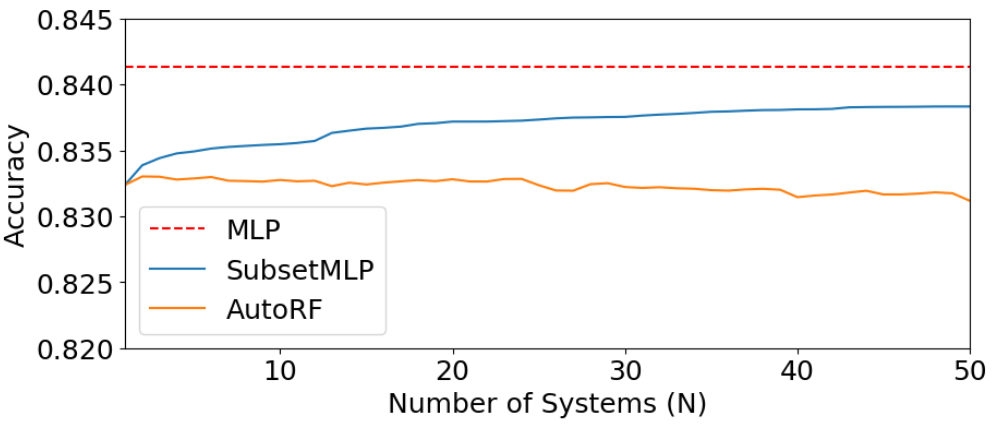}
  \end{center}
  \caption{All accuracies are computed using cross-validation. The top line "MLP" is obtained by selecting the best (out of 600) Student per task. The "SubsetMLP" and "AutoRF" curves show the performance of selecting from a subset of size N.}
  \label{fig:Auto_RF}
\end{figure}

As input features to this system, we used metadata characterizing the datasets as provided by OpenML (\cite{OpenmlMeasure}). 
We excluded features corresponding to performance of other reference classifiers, e.g. nearest neighbor.
As a result, our input feature vectors were formed by 74 coefficients reflecting various dataset qualities and quantities. 
Fig.~\ref{fig:Auto_RF} shows the comparison of choice of the best candidate from the set of a particular size with the automatic choice done by the random forest.
We conclude that the automatic Student selection fails to select the best Student candidates.
Our reasoning is that the metadata provided by OpenML for dataset characterization are not suitable for automatic system selection and the performance is also affected by the small number of samples available.

\section{Conclusions and Future Work}
\label{sec:conclusions}

There are multiple benefits in being able to represent machine learning pipelines for various datasets and tasks in a unified framework based on neural networks.
In this work, we first laid out a solution for conversion of generic machine learning pipelines into neural networks.
We view the conversion as a multi-stage process where parts of the original pipeline are first converted separately before joint optimization can be done.
We then focused on the conversion of the back-end classifier represented by random forest into a NN.
We showed that NNs learned employing the student-teacher concept performed generally as well as the original random forests, with a few outliers.
While NNs with many different configurations were initially considered, we showed that the number of NN configurations can be significantly reduced without harming the performance.

Finally, we investigated the possibility of using a random forest for automatic selection of the best NN configuration based on the characteristics of the data.
In contrast to using a single best configuration, this automatic selection leads to only a marginal improvement for very small sets of Students and the performance deteriorates as the number of Students grows.
We attribute this mainly to the lack of relevant information in the metadata which is used as input to the automatic selection system, and the lack of training samples.

Experimental work presented in this paper represents just an initial step in our effort and many aspects of our proposed solution will have to be further investigated in the future, such as substitution of various parts of generic ML pipelines (including feature extraction or transformation), augmentation of training data, end-to-end joint optimization and automatic selection of the best NN configuration for substitution.

\section*{Acknowledgement}
This work is sponsored by the Air Force Research Laboratory (AFRL) and DARPA.

\bibliography{ref}
\end{document}